\documentclass[letterpaper, 10 pt, conference]{ieeeconf}

\usepackage[utf8]{inputenc} 
\usepackage[T1]{fontenc}    
\usepackage{cite}
\makeatletter
\let\NAT@parse\undefined
\makeatother
\usepackage{hyperref}
\usepackage{url}            
\usepackage{booktabs}       
\usepackage{amsfonts}       
\usepackage{nicefrac}       
\usepackage{microtype}      
\usepackage{xcolor}         
\usepackage{amsmath}
\usepackage{amssymb}
\usepackage{logicpuzzle}
\usepackage{stmaryrd}
\usepackage{algorithmic}
\usepackage{algorithm}
\usepackage{makecell}

\usepackage{amsmath,amsfonts}
\usepackage{algorithmic}
\usepackage{algorithm}
\usepackage{array}
\usepackage[caption=false, font=footnotesize]{subfig}
\usepackage{textcomp}
\usepackage{stfloats}
\usepackage{url}
\usepackage{verbatim}
\usepackage{graphicx}
\usepackage{epstopdf}
\usepackage[utf8]{inputenc}
\usepackage[T1]{fontenc}
\usepackage{booktabs} 
\usepackage{diagbox}  
\usepackage{multirow} 
\usepackage{threeparttable} 
\usepackage{epstopdf} 
\usepackage{colortbl}  
\usepackage{hhline}
\usepackage[inkscapelatex=false]{svg}
\definecolor{codegreen}{rgb}{0,0.5,0}
\definecolor{codeblue}{rgb}{0,0,0.9}
\definecolor{codeblues}{rgb}{0,0,0.4}
\definecolor{codegray2}{rgb}{0.4,0.4,0.4}
\definecolor{codegray}{rgb}{0.9,0.9,0.9}

\usepackage{color, xcolor}

\newcounter{ALC@tempcntr}
\newcommand{\LCOMMENT}[1]{%
    \setcounter{ALC@tempcntr}{\arabic{ALC@rem}}
    \setcounter{ALC@rem}{1}
    \item //#1 
    \setcounter{ALC@rem}{\arabic{ALC@tempcntr}}
}%

\usepackage{fontawesome}
\definecolor{Dred}{HTML}{C81D31}
\definecolor{Dgreen}{HTML}{588E31}
\definecolor{codegray}{rgb}{0.9,0.9,0.9}

\IEEEoverridecommandlockouts

\begin{document}
\title{DAG-Plan: Generating Directed Acyclic Dependency Graphs for Dual-Arm Cooperative Planning}

\author{Zeyu Gao$^{1*}$, Yao Mu$^{2*}$, Jinye Qu$^{1}$, Mengkang Hu$^{4}$, Shijia Peng$^{5}$, Chengkai Hou$^{3}$, Lingyue Guo$^{1}$, 
\quad
\\
Ping Luo$^{4,5}$, Shanghang Zhang$^{3\dagger}$ and Yanfeng Lu$^{1\dagger}$
\thanks{
Zeyu Gao* and Yao Mu* are co-first authors. Yanfeng Lu$^{\dagger}$ and Shanghang Zhang$^{\dagger}$ are the corresponding authors. This work is supported by the Strategic Priority Research Program of the Chinese Academy of Sciences (XDA0450200, XDA0450202), and Central Government Guidance Funds for Local Science and Technology Development (YDZX2025124), the National Natural Science Foundation of China (62476011), and the Beijing Natural Science Foundation (L252060).}
\thanks{
$^1$ Zeyu Gao, Jinye Qu, Lingyue Guo and Yanfeng Lu are with the State Key Laboratory of Multimodal Artificial Intelligence Systems, Institute of Automation, Chinese Academy of Sciences (CASIA), Beijing 100190, China, and also with the School of Artificial Intelligence, University of Chinese Academy of Sciences (UCAS), Beijing 100049, China (e-mail: gaozeyu2023@ia.ac.cn; yanfeng.lv@ia.ac.cn).}
\thanks{
$^2$ Yao Mu is with the School of Computer Science, Shanghai Jiao Tong University, Shanghai 200240, China (e-mail: muyao@sjtu.edu.cn)}
\thanks{
$^3$ Chengkai Hou and Shanghang Zhang are with State Key Laboratory of Multimedia Information Processing, School of Computer Science, Peking University, Beijing 100871, China (e-mail: shanghang@pku.edu.cn).}
\thanks{
$^4$ Mengkang Hu and Ping Luo are with the Department of Computer Science, The University of Hong Kong, Hong Kong 999077, China}
\thanks{
$^5$ Shijia Peng and Ping Luo are with OpenGVLab, Shanghai AI Laboratory, Shanghai 200232, China}
}

\maketitle

\begin{abstract}

Dual-arm robots promise greater efficiency but require planning for complex tasks with nonlinear sub-task dependencies. Current methods using Large Language Models (LLMs) suffer from a fundamental trade-off: generating linear sequences is efficient but fails to model parallelism and adapt to changes, while iterative querying is adaptive but too slow and costly.
To bridge this gap, we introduce DAG-Plan, a novel task planning framework that for the first time employs a Directed Acyclic Graph (DAG) as the central representation for dual-arm coordination. The key insight is that a DAG natively captures complex sub-task dependencies and explicitly reveals opportunities for parallel execution. Within this framework, an LLM is used only once as a powerful semantic parser to translate a natural language instruction into a structured DAG. During execution, our system dynamically assigns candidate nodes to the suitable arm based on real-time environmental observations, enabling truly adaptive and parallel operation.
Extensive evaluation on a dual-arm kitchen benchmark shows that DAG-Plan's structured approach fundamentally outperforms existing paradigms. It achieves a 48\% higher success rate than single-query linear sequence methods with dual arm by robustly managing dependencies, and an 84.1\% higher execution efficiency than iterative querying methods by eliminating the latency of repeated LLM calls. Our work demonstrates that a principled, graph-based representation is the key to unlocking efficient and reliable LLM-based planning for complex robotic systems.
More demos and code are available on \href{https://sites.google.com/view/dag-plan}{https://sites.google.com/view/dag-plan}.

\end{abstract}


\section{Introduction}
Achieving effective bimanual coordination in robotics is challenging due to the complexities and long-horizon of dual-arm operations, requiring precise spatial and temporal coordination~\cite{ying2021deep,borrell2024optimization}. While humans effortlessly coordinate their hands in daily long-horizon tasks, replicating such coordination in robots presents significant challenges. Traditional method employed hand-designed primitives to manage the movements of dual robotic arms~\cite{mirrazavi2017dynamical,grannen2022learning}. Whereas these methods often fall short in long-horizon task as they lack the flexibility required for adaptive task execution. Large language models (LLMs) have emerged as powerful tools endowed with extensive knowledge and sophisticated reasoning abilities. By systematically breaking down tasks into actionable sub-tasks and leveraging their commonsense knowledge and implicit reasoning capabilities, LLMs empower robots to effectively adapt to long-horizon tasks in the wild~\cite{mu2023embodied,mu2024robocodex}.

The prevalence of linear task sequences, however, introduces a fundamental bottleneck for dual-arm coordination. While LLM-based planners have demonstrated remarkable reasoning capabilities, we observe that their performance in bimanual scenarios is constrained by the inherent limitations of the linear representation paradigm itself. This is true even for advanced multi-agent frameworks ~\cite{singh2024twostep, mandi2024roco, zhao2025dual} that decompose tasks for multiple robots: their output remains a set of linear steps, which fails to explicitly model the complex, non-linear dependencies and parallel opportunities critical for efficient dual-arm execution. Consequently, planners are forced to choose between the rigidity of a single, pre-generated linear sequence and the prohibitive inefficiency of iterative LLM querying to navigate execution.

We identify that these challenges are not independent issues but rather symptomatic manifestations of a single root cause: the reliance on linear task representations. This representation is fundamentally mismatched to the problem for three key reasons. First, its sequential nature is incapable of explicitly representing the parallel execution opportunities that are essential for efficiency. Second, a pre-generated linear sequence is inherently rigid, decoupled from the real-time state of the environment and unable to adapt the execution order or arm assignment dynamically. Third, while iterative linear planning attempts to recover adaptability by frequently querying the LLM, it does so at the prohibitive cost of latency and computational expense, merely treating the symptom rather than curing the disease. Thus, the field is currently trapped in a false choice between inflexibility and inefficiency.

To break this deadlock, we propose DAG-Plan, a novel framework that replaces the linear representation with a structured Directed Acyclic Graph (DAG). DAG explicitly captures sub-tasks as nodes and, crucially, their temporal dependencies as directed edges, thereby natively encoding both the execution constraints and the parallel opportunities that are opaque to a linear sequence. Within this framework, the role of the LLM is refined to a single, initial query where it acts as a powerful semantic parser, translating the natural language instruction into this structured DAG. The subsequent planning and execution intelligence resides in our Task Planning Inference module. This module continuously monitors the environmental state and the DAG's structure—dynamically assigning the feasible and efficient pair of executable sub-tasks to the dual arms. This enables true parallel and adaptive execution without the latency of repeated LLM queries.
Our main contributions are summarized as follows:
\begin{itemize}
\item We identify that the bottleneck in LLM-based dual-arm planning is the reliance on linear task representations, which are ill-suited for expressing the complex, parallelizable nature of bimanual tasks and necessitate a trade-off between efficiency and adaptability.
\item We propose DAG-Plan, a novel framework that introduces a Directed Acyclic Graph (DAG) as the structured task representation for dual-arm planning. This representation explicitly models sub-task dependencies and parallelism, enabling efficient and adaptive execution with a single LLM query.
\item We demonstrate that our approach outperforms existing paradigms. It achieves a 48\% higher success rate than single-query linear sequence methods with dual arm by robustly managing dependencies, and an 84.1\% higher execution efficiency than iterative querying methods by eliminating the latency of repeated LLM calls.


\end{itemize}

\section{Related Works}

\subsection{Task Planning with LLMs} 

Large Language Models have revolutionized task planning by translating natural language goals into executable action sequences for embodied agents~\cite{brohan2023can, hu2023tree}. Their success stems from an unparalleled ability to leverage commonsense knowledge and in-context learning. However, a prevalent limitation across this field is the reliance on linear action sequences as the primary output representation.
This limitation becomes acutely apparent in multi-entity systems, such as multi-robot teams~\cite{singh2024twostep, mandi2024roco} or dual-arm robots~\cite{liu2023llm,joublin2023copal,zhao2025dual}. While these works demonstrate advanced decomposition or communication strategies (e.g., independent task decomposition in Twostep~\cite{singh2024twostep} or inter-robot discussion in RoCo~\cite{mandi2024roco}), their final output remains a linear sequence of steps for each agent. This representation obscures the temporal dependencies and parallel opportunities between sub-tasks, forcing a trade-off between plan quality and execution efficiency. 

\subsection{Dual-arm Robot Manipulation} 

Research on dual-arm manipulation has made significant strides, primarily at the skill level, across industrial~\cite{ying2021deep,borrell2024optimization}, agricultural~\cite{sepulveda2020robotic,yoshida2022automated}, and domestic settings~\cite{ogren2012multi,ju2023kinematic}. This progress is driven by advances in motion planning~\cite{ju2023kinematic}, foundation models~\cite{fang2023anygrasp}, and learning-based methods~\cite{jiang2023mastering, fu2024mobile}, enabling robots to perform human-like bimanual skills such as lifting or assembling.
However, autonomously orchestrating a long-horizon task composed of multiple such skills remains a open challenge. While LLMs have been applied to dual-arm systems, they often only control one arm at a time~\cite{liu2023llm} or rely on pre-defined bimanual skills~\cite{joublin2023copal}, thus failing to fully exploit the system's potential for parallel and adaptive operation at the task level. Our work addresses this gap by proposing a framework for task-level planning that dynamically composes and coordinates skills for both arms.

\subsection{Structured Task Decomposition (STD)} 

STD aims to break down complex tasks into structured representations like graphs, explicitly modeling dependencies and parallelism. Traditional STD methods were heavily constrained by data availability, relying on crowd-sourcing~\cite{kokkalis2013taskgenies,zhou2022show} or complex query systems~\cite{hassan2014supporting,mehrotra2017extracting}. The advent of LLMs, with their vast repositories of commonsense knowledge, has emerged as a powerful new paradigm for STD. For instance, TaskLAMA~\cite{yuan2024tasklama} demonstrated that LLMs can effectively decompose tasks into graphs with temporal relations. This line of work provides a critical foundation, confirming the feasibility of using LLMs for structured output. DAG-Plan pushes this frontier further. We explore how to leverage LLM-generated DAG not merely as a structured output, but as the central representation for real-time, adaptive, and parallel execution on a dual-arm robotic system.

\subsection{Comparison with Symbolic Planners}
Our approach is also related to symbolic planners like PDDL~\cite{aeronautiques1998pddl} and Scene Graph~\cite{armeni20193d}, which offer formal guarantees but require hand-crafted world models, creating a significant scalability bottleneck.
PDDL and Scene Graph offers high formalism and precision, but low generalizability, and require significant human effort to define the domain.
Our DAG-Plan method offers high generalizability and requires minimal human effort only a skill and task nature language description but provides less formal guarantee and relies on the LLM's implicit knowledge.
This makes DAG-Plan suited for open-world environments where formal models are impractical, while still producing a more explicit and parallelizable plan than a linear sequence.

\section{Methods}

\begin{figure*}[t]
    \vspace{3pt}
      \centering
\includegraphics[width=0.89\linewidth]{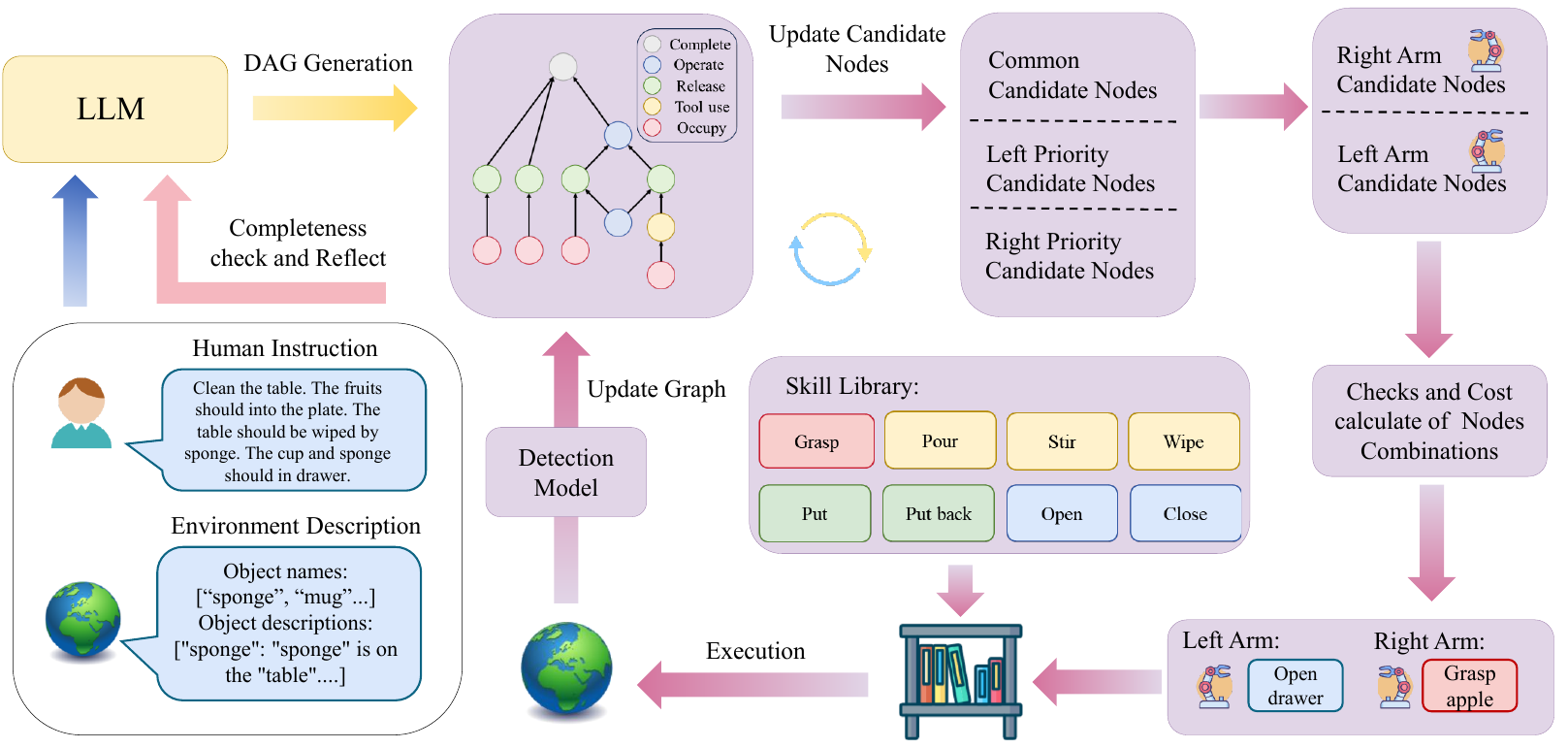}
      \vspace{-5pt}
      \caption{An overview of DAG-Plan. The DAG-Plan generates a DAG based on human instruction and environmental description. It checks the graph's completeness and reflects the LLM to regenerate if incomplete. Once a valid DAG is obtained, DAG-Plan performs task inference to identify executable candidate nodes. The occupied arm and free arm are assigned priority candidate nodes and common candidate nodes respectively. The framework then evaluates all candidate combinations for feasibility and cost. DAG-Plan selects the nodes with the lowest cost and employs skill in library for execution. DAG-Plan updates the graph, iterating inference until the DAG is fully executed.}
      \vspace{-10pt}
      \label{framework}
\end{figure*}

The DAG-Plan framework is designed to overcome the fundamental limitations of linear representations by introducing a structured DAG as the core planning representation. This representation natively captures complex temporal dependencies and explicit parallelism, enabling efficient and adaptive dual-arm coordination with a single LLM query. An overview of DAG-Plan pipeline is illustrated in Fig. \ref{framework}.

\subsection{Sub-task Dependency Directed Acyclic Graph Generation}

We propose a novel approach that uses an LLM as a semantic parser to decompose a complex natural language instruction into a structured Directed Acyclic Graph (DAG). This graph representation is fundamental to our framework, as it natively captures the temporal dependencies and parallel opportunities inherent in bimanual tasks, which are opaque to linear sequences.

We formally define the dual-arm task graph as a tuple:
\begin{equation}
G = (V, E, T, N),
\end{equation}
\noindent where:
\begin{itemize}
    \item $V = \{v_1, v_2, ..., v_n\}$ is the set of sub-task nodes.
    \item $E \subseteq V \times V$ is the set of directed edges representing temporal dependencies. An edge $e_{ij} = (v_i, v_j) \in E$ indicates that sub-task $v_i$ must be completed before $v_j$ can begin.
    \item $T: V \to \mathcal{T}$ maps each node to a task type from the set $\mathcal{T}$.
    \item $N: V \to \{1, 2\}$ specifies the number of arms required for the execution of each node.
\end{itemize}

The task type set $\mathcal{T}$ is defined as \{\textit{Occupy}, \textit{Tool use}, \textit{Release}, \textit{Operate}, \textit{Complete}\}. This typology is critical for modeling arm state transitions and ensuring logical soundness:
\begin{itemize}
    \item \textit{Occupy}: Engages a gripper, leaving the arm \textit{occupied} (e.g., grasp).
    \item \textit{Tool use}: Requires a tool, using it and keeping it in the gripper (e.g., wipe).
    \item \textit{Release}: Releases an object, freeing the arm (e.g., place).
    \item \textit{Operate}: A general task that occupies the arm only during execution (e.g., open).
    \item \textit{Complete}: The terminal node means the task is finished when DAG only remain this node.
\end{itemize}

The task type $T(v)$ dictates the state transition of the robotic arms. We define an \textit{occupy-release} pair as a critical subgraph $G' \subseteq G$ that manages object manipulation:
\begin{equation}
\begin{aligned}
&G' = (V', E'),\text{ s.t. } V' = \{v_o, v_{t1}, ..., v_{tk}, v_r\} \\
T(v_o)& = \textit{Occupy}, \quad T(v_r) = \textit{Release}, \quad T(v_t) = \textit{Tool use},\\
\end{aligned}
\end{equation}
where a path $v_o \leadsto v_r$ exists via $E'$.
This structure enforces the fundamental constraint that for any release operation, there must exist a corresponding, preceding occupy operation, ensuring an object is grasped before it is released.

The LLM is prompted to generate a DAG. The output is validated for structural integrity. Graphs with incomplete \textit{occupy-release} pairs or disconnected components are rejected and regenerated, ensuring a complete plan for execution.

\subsection{Task Planning Inference with Directed Acyclic Graph}
The generated DAG serves as a static plan. The Task Planning Inference module is the dynamic execution engine that traverses this graph in real-time, enabling adaptive and parallel operation. Its core function is to continuously select the optimal pair of executable sub-tasks for the two arms based on the current graph state and environmental observations.

\subsubsection{Stateful Node Selection}
The executor maintains the state of the task graph. Each node $v_i \in V$ has an execution status $s(v_i) \in \{\textit{Pending}, \textit{Ready}, \textit{Done}\}$. A node becomes \textit{Ready} when all its prerequisite nodes are \textit{Done}.

The set of executable nodes is dynamically prioritized based on the semantic context of the graph:
\begin{itemize}
    \item \textit{Common Candidates} ($C_c$): The set of all \textit{Ready} nodes.
    \item \textit{Priority Candidates} ($C_p$): For any \textit{Ready} node $v_r$ where $T(v_r) = \textit{Release}$, its entire antecedent path within the same \textit{occupy-release} pair is given priority. This ensures that arms committed to an ongoing manipulation sequence complete it before starting new ones, which is critical for preventing logical errors and deadlocks. Each left and right arm has a Priority Candidates 
\end{itemize}

\subsubsection{Execution Loop}
The executor runs a continuous loop:
First, update candidate nodes: Recompute $C_c$ and identify $C_p$ based on the new graph state.
Second, checks and cost calculate of nodes combinations: For each arm, if a priority candidate exists, nodes are selected from them. Otherwise, nodes are selected from the common candidates. 
Third, update graph: Upon a node's completion, it is marked \textit{Done} and removed from $G$. This updates the \textit{Ready} status of its successors.

\subsubsection{Feasibility Checking and Cost Calculation}
The system evaluates all possible pairs of selected nodes. Each candidate assignment is validated through a series of geometric feasibility checks:
\begin{itemize}
    \item \textit{Dependency Check}: A logical check rejects a candidate if it is an \textit{Occupy} node whose subsequent \textit{Release} node is not yet \textit{Ready}, preventing an arm from being indefinitely occupied.
    \item \textit{Reachable Check}: Calculates the Euclidean distance between target objects for both arms. A pair is rejected if distance between left target and right target is greater than $d_{\text{Reachable}}$, ensuring both targets are within a feasible simultaneous workspace.
    \item \textit{Across Avoidance}: The distance between the two target object must greater than a safety threshold $d_{\text{Across}}$, indicating a high risk of inter-arm collision. This prevents scenarios where the arms would need to cross each other or operate in dangerously close proximity.
\end{itemize}

Object position for these checks are acquired using Grounding DINO~\cite{liu2024grounding} and SAM2~\cite{ravi2024sam} for segmentation, and SAR-Net~\cite{lin2022sar} for 6D pose estimation of specific categories.
After completing the checks, we calculate the distance to left and right hand of the candidate nodes based on the environment state as the cost. We aim for the target objects to be close to the robotic hands facilitating dual-arm operations. The pair with the lowest combined cost is selected for execution. Finally, we briefly introduce the task planning inference process in Alg. \ref{dag_plan_algorithm}.

\begin{figure*}[t]
    \vspace{3pt}
      \centering
   \includegraphics[scale=0.46]{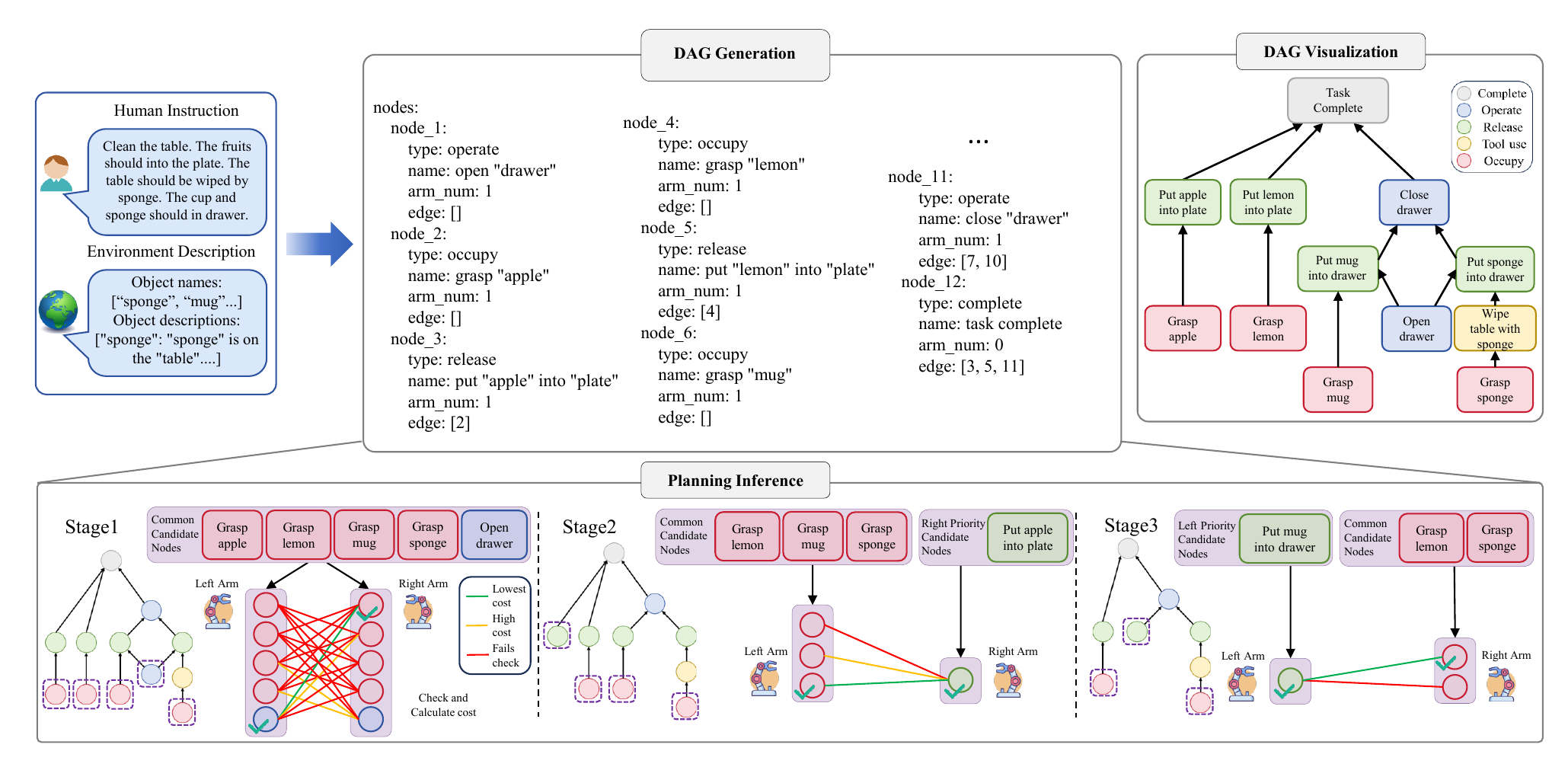}
   \vspace{-10pt}
      \caption{The process of Task Planning Inference. In task 2 ``clean the table (Hard)'', DAG-Plan initializes common candidate nodes based on the DAG. It evaluates node combinations, checks feasibility, and calculates costs. In stage 1, the right arm is selected to grasp apple and the left to open drawer. After execution, the task graph and nodes are updated, adding subsequent \textit{release} nodes to the priority candidate nodes for right arm. In stage 2, right arm is assigned corresponding priority candidate nodes, checked and left arm still selected node in common candidate nodes with empty priority candidate nodes. The task graph and nodes are updated again. In stage 3, left arm put mug into drawer and right arm grasp lemon.}
      \label{planning}
     \vspace{-8pt}
\end{figure*}

\begin{algorithm}
\caption{Task Planning Inference with DAG}
\label{dag_plan_algorithm}
\begin{algorithmic}[1]
\STATE \textbf{Input:} Task Graph $G = (V, E, T, N)$
\STATE \textbf{Initialize:} $s(v_i) \gets \text{Pending}, \forall v_i \in V$
\WHILE{$\exists v_i \in V \mid s(v_i) \neq \text{Done} \land T(v_i) \neq \text{Complete}$ }
    \LCOMMENT{ Update candidate nodes }
    \STATE $C_c \gets \{v_i \in V \mid s(v_i) = \text{Ready}\}$
    \STATE $C_{pl} \gets \{v \in \text{left occupy-release paths} \mid v \in C_c\}$
    \STATE $C_{pr} \gets \{v \in \text{right occupy-release paths} \mid v \in C_c\}$
    \STATE $C_{l} \gets (C_{pl} \neq \emptyset) ? C_{pl} : C_c$
    \STATE $C_{r} \gets (C_{pr} \neq \emptyset) ? C_{pr} : C_c$
    \LCOMMENT{ Checks and cost calculate of nodes combinations }
    \STATE Filter feasible pairs $(v_l, v_r) \in C_{l} \times C_{r}$ through dependency, reachability, and across-arm checks
    \STATE Calculate cost for each feasible pair based on distance to hands $(v_l^*, v_r^*) \gets \arg\min \text{cost}(v_l, v_r)$
    \STATE $\text{Execute}(v_l^*, v_r^*)$
    \LCOMMENT{ Update graph }
    \STATE $s(v_l^*) \gets \text{Done}$, $s(v_r^*) \gets \text{Done}$
    \FOR{$v_i \in V \mid s(v_i) = \text{Pending}$}
        \IF{$\forall (v_j, v_i) \in E, s(v_j) = \text{Done}$} 
            \STATE $s(v_i) \gets \text{Ready}$
        \ENDIF
    \ENDFOR
\ENDWHILE
\end{algorithmic}
\end{algorithm}

\subsection{Sub-tasks Execution with Foundation Model}

Once the sub-tasks to be executed are determined, the robot needs to perform the corresponding actions to bridge the gap between textual instructions and the physical environment. The collective perceptual inferences specific to objects, along with physical insights and parameters for manipulation, are methodically organized and converted into a structured format of executable action code.

\subsubsection{Occupy}
We use the pre-trained AnyGrasp\cite{fang2023anygrasp} to produce a variety of grasp pose proposals. 
\subsubsection{Tool use}
We have specifically designed tool-usage skills based on the categories of tools employed, and have provided adjustable parameters to enable the robot to perform flexible tool-manipulating actions.
\subsubsection{Release}
Based on the current grasping position of the object and the 3D bounding box of the target object, we propose an appropriate release pose to ensure that the target object can be accurately placed at the desired location.
\subsubsection{Operate} 
We employ the GAPartNet\cite{geng2023gapartnet} to forecast the physical characteristics of articulated objects. This model works by dividing the point cloud of an articulated object into its constituent rigid parts and subsequently calculating the articulation parameters.


\begin{figure*}[t!]  
    \vspace{5pt}
    \centering
    \includegraphics[width=0.9\linewidth]{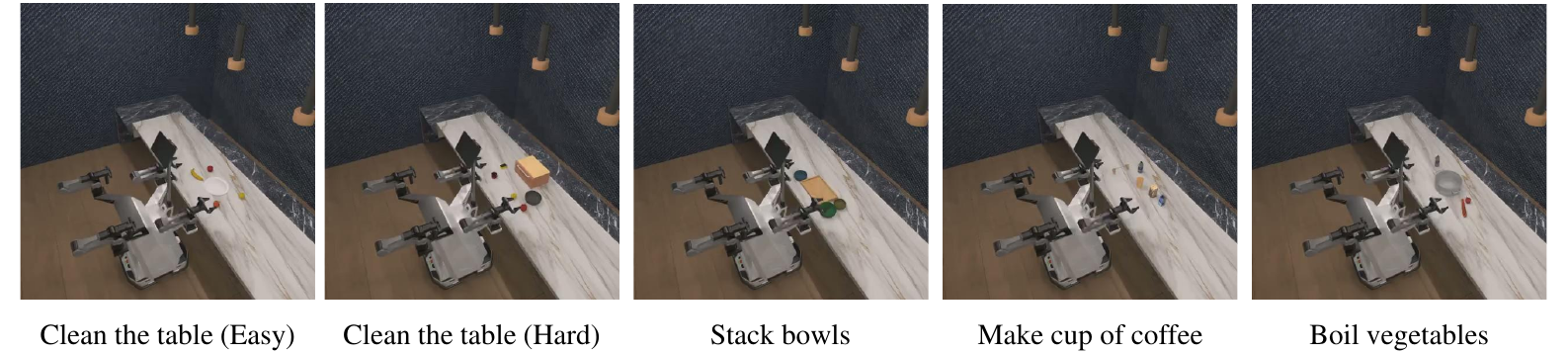}
    \vspace{-5pt}
    \caption{Snapshots of 5 Tasks of Dual-arm Kitchen Benchmark.}   
    \label{kitchen}
\end{figure*}

\begin{table*}[t!]
\caption{Task list of Dual-arm Kitchen Benchmark.}
    \vspace{-5pt}
    \centering
    \vspace{-5pt}
    \footnotesize
    \resizebox{0.95\linewidth}{!}{
\begin{tabular}{c|l|l}
\toprule
\textbf{Index} & \textbf{Task Name} & \textbf{Human Instruction} \\
\midrule
Task 1 & Clean the table (Easy) & Clean the table. Put objects into plate. \\

Task 2 & Clean the table (Hard) & Clean the table. The fruits should into the plate. The table should be wiped by sponge. The mug and sponge should in drawer.\\ 

Task 3 & Stack bowls & Stack the bowls onto the wooden tray with the green bowl, blue bowl and yellow bowl order.\\

Task 4 & Make cup of coffee & Make a cup of coffee. You should add the coffee, water and milk in order. Finally, stir it with the spoon. \\

Task 5 & Boil vegetables &
Boil vegetables. Pour water and put vegetables into the pot.\\
\bottomrule
\end{tabular}
}

\vspace{-10pt}
\label{tab:task list}
\end{table*}

\section{Experiments}
 In the experiments, we are interested in answering the following questions about DAG-Plan:
\begin{itemize}
\item[$\bullet$] \textbf{Q1}: To what extent can the task-graph approach improve performance compared to the task-sequence method?
\item[$\bullet$] \textbf{Q2}: How does the cost and efficiency of DAG-Plan compare to iterative methods?
\item[$\bullet$] \textbf{Q3}: How does the performance of DAG-Plan compare to multi-agents methods?
\item[$\bullet$] \textbf{Q4}: Can DAG-Plan be deployed in real-world setting?
\end{itemize}

\subsection{Experimental Setup}
\label{sec: Experimental Setup}

To validate the correctness and execution efficiency of our method, we created a Dual-arm Kitchen Benchmark, including plan tests and physical simulation tests. This benchmark fills the gap in long-sequence operation physics simulation benchmarks for dual-arm robots. The goal of this benchmark is to validate the success rate and efficiency of dual-arm robot planning and executable ability in complex scenarios. The dual-arm robot should correctly plan and fully utilize both the left and right arms, completing tasks with as few execution stages as possible. The benchmark consists of 5 sequential tasks are shown in Tab. \ref{tab:task list} and Fig. \ref{kitchen}, comprising a total of 44 sub-tasks. Our physical simulation scene is built on the Sapien~\cite{xiang2020sapien}. The embodied platform is Agilex CobotMagic, each 6 DOF arm equipped with a two-finger gripper. The platform is equipped with an Intel RealSense L515 RGB-D camera, attached on the robot's head.

\subsubsection{Evaluation of Planning Effectiveness and Conciseness} In this experiment, we focus on testing the conciseness of the generated plans and the number of stages required. In a stage, the robot can execute a right arm node and a left arm node. This requires that the plans generated by the LLMs can achieve the task goals in terms of language logic and do not violate the preconditions for stage execution. 
We used GPT-4o to generate 10 plans for each task, evaluating their \textbf{Success Rate (SR)} and the minimum \textbf{Stage} of the passed plan required at language level. Due to cost considerations, we also include average \textbf{Tokens} usage as an important metric for evaluating cost of the algorithm. Finally, we calculated the average success rate and the average number of stages for all tasks. We defined \textbf{Stage Efficiency} as the ratio of single-arm plan stages to the stages required by each method. For failed dual-arm plans, we calculated the stage count based on the single-arm plan stages to ensure a fair comparison.

\subsubsection{Evaluation with Physical Simulation} In this experiment, we will test the executability and execution efficiency of the plans in physical simulation scenarios. Compared to plan tests, physical simulation tests validate both the high-level planning and low-level execution capabilities. Then evaluate the \textbf{Success Rate (SR)} and minimum \textbf{Time} which consists of \textcolor{Dgreen}{query time} and \textcolor{Dred}{execution time} in the simulator with 10 trials. Finally, we calculated the average success rate and the time for all tasks. We defined \textbf{Execution Efficiency} as the ratio of single-arm plan time to the time required by each method. For failed dual-arm plans, we calculated the average time based on the single-arm time to ensure a fair comparison.

\subsubsection{Baseline Algorithms and Proposed Method} 
We compare the planning algorithms baselines and our proposed method:
\begin{enumerate}
\item{\textit{Task Planning for Single-arm (TP-S)}:} TP-S directly uses LLMs to generate a full task sequence, with each stage involving a single arm to manipulate a single object.
\item{\textit{Task Planning for Dual-arm (TP-D)}:} TP-D directly uses LLMs to generate a full task sequence, with each stage involving dual arms.
\item{\textit{Twostep}~\cite{singh2024twostep}:} Twostep decomposes complex tasks into fully independent sub-tasks with dual agents, with the primary agent and the secondary agent each completing separate sub-tasks to accelerate task execution.
\item{\textit{RoCo}~\cite{mandi2024roco}:} RoCo proposed multi-agents collaboration using LLMs for planning which is an iterative method. Robots query LLMs in each stages to collaboratively reason about task strategies. DABICO~\cite{zhao2025dual} is a dual-armed version of the RoCo, and is equivalent to the RoCo in our experiments settings.
\item{\textit{DAG-Plan (Proposed)}:} Our method DAG-Plan generates a task graph, followed by task planning inference to iteratively generate nodes for each stage.
\end{enumerate}

\subsection{Evaluation of Planning Effectiveness and Conciseness}
\label{sec:Results}

\begin{figure*}[thpb]
      \vspace{3pt}
      \centering
      \includegraphics[scale=0.53]{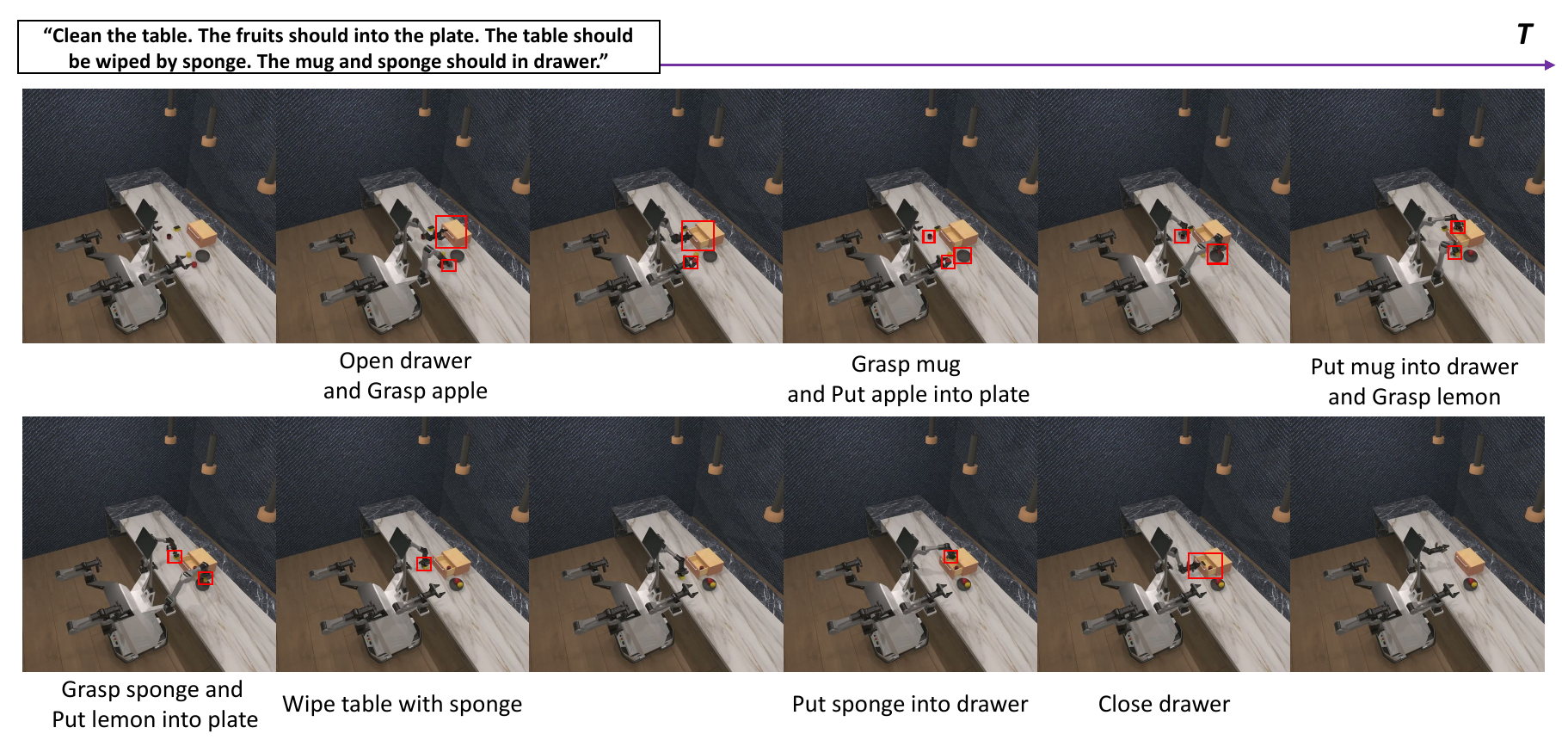}
      \vspace{-10pt}
      \caption{Simulation snapshots of the execution process of long-horizon task 2.}
      \vspace{-5pt}
      \label{demo}
\end{figure*}

\begin{table*}[htbp]
  \footnotesize
  \begin{center}
    \caption{Performance comparison on plan tests. We report the success rate, minimum stage and average token usage of the 10 plans generated by the LLM for each task.}
    \vspace{-3pt}
    \label{plan_test_results}
     \resizebox{1.0\linewidth}{!}{
    \begin{tabular}{l|c|c|c|c|c|c|c|c|c|c|c|c|c|c|c|c|c|c}
    \toprule
      \textbf{} & \multicolumn{3}{c|}{\textbf{Task1}} & \multicolumn{3}{c|}{\textbf{Task2}} & \multicolumn{3}{c|}{\textbf{Task3}} & \multicolumn{3}{c|}{\textbf{Task4}} & \multicolumn{3}{c|}{\textbf{Task5}} & \multicolumn{3}{c}{\textbf{Average}} \\
      \textbf{} & \textbf{SR}$\uparrow$ & \textbf{Stage}$\downarrow$ & \textbf{Tokens}$\downarrow$ & \textbf{SR}$\uparrow$ & \textbf{Stage}$\downarrow$ & \textbf{Tokens}$\downarrow$ & \textbf{SR}$\uparrow$ & \textbf{Stage}$\downarrow$ & \textbf{Tokens}$\downarrow$ & \textbf{SR}$\uparrow$ & \textbf{Stage}$\downarrow$ & \textbf{Tokens}$\downarrow$ & \textbf{SR}$\uparrow$ & \textbf{Stage}$\downarrow$ & \textbf{Tokens}$\downarrow$ & \textbf{SR}$\uparrow$ & \textbf{Stage Efficiency}$\uparrow$ & \textbf{Tokens}$\downarrow$\\
      \midrule
      \textbf{TP-S} & $100\%$ & $8$ & $2030.0$ & $70\%$ & $11$ & $2424.8$ & $100\%$ & $6$ & $2086.8$ & $90\%$ & $12$ & $2413.0$ & $100\%$ & $7$ & $2096.6$ & $92\%$ & $100.0\%$ & $2210.2$ \\
      \textbf{TP-D} & $100\%$ & $4$ & $2084.1$ & $0\%$ & $Fail$ & $2355.6$ & $0\%$ & $Fail$ & $2038.6$ & $20\%$ & $8$ & $2405.2$ & $80\%$ & $5$ & $2040.6$ & $40\%$ & $129.4\%$ & $2180.8$\\
      \textbf{TwoStep\cite{singh2024twostep}} & $100\%$ & $4$ & $3699.6$ & $0\%$ & $Fail$ & $4248.2$ & $0\%$ & $Fail$ & $3823.4$ & $0\%$ & $Fail$ & $4130.2$ & $100\%$ & $5$ & $3810.2$ & $40\%$ & $115.8\%$ & $3942.3$\\
      \textbf{RoCo\cite{mandi2024roco}} & $100\%$ & $4$ & $26615.0$ & $0\%$ & $Fail$ & $68921.4$ & $100\%$ & $4$ & $25410.2$ & $40\%$ & $8$ & $65242.0$ & $100\%$ & $4$ & $26985.0$ & $68\%$ & $141.9\%$ & $42634.7$\\
      \cellcolor{codegray}\textbf{DAG-Plan} & \cellcolor{codegray}$100\%$ & \cellcolor{codegray}$4$ & \cellcolor{codegray}$2031.0$ & \cellcolor{codegray}$70\%$ & \cellcolor{codegray}$7$ & \cellcolor{codegray}$2207.2$ & \cellcolor{codegray}$100\%$ & \cellcolor{codegray}$4$ & \cellcolor{codegray}$1980.6$ & \cellcolor{codegray}$70\%$ & \cellcolor{codegray}$7$ & \cellcolor{codegray}$2277.0$ & \cellcolor{codegray}$100\%$ & \cellcolor{codegray}$4$ & \cellcolor{codegray}$1973.0$ & \cellcolor{codegray}$88\%$ & \cellcolor{codegray}$169.2\%$ & \cellcolor{codegray}$2093.8$\\
    \bottomrule
    \end{tabular}
    }
  \end{center}
  \vspace{-13pt}
\end{table*}

\begin{table*}[htbp]
\footnotesize
  \begin{center}
    \caption{Performance comparison on physical simulation tests. We report the success rate and minimum time for each task with 10 trials. Time consists of \textcolor{Dgreen}{query time} and \textcolor{Dred}{execution time}.}
    \vspace{-3pt}
    \label{Physical test}
    \resizebox{1.0\linewidth}{!}{
    \begin{tabular}{l|c|c|c|c|c|c|c|c|c|c|c|c}
    \toprule
      \textbf{} & \multicolumn{2}{c|}{\textbf{Task1}} & \multicolumn{2}{c|}{\textbf{Task2}} & \multicolumn{2}{c|}{\textbf{Task3}} & \multicolumn{2}{c|}{\textbf{Task4}} & \multicolumn{2}{c|}{\textbf{Task5}} & \multicolumn{2}{c}{\textbf{Average}}\\
      \textbf{} & \textbf{SR}$\uparrow$ & \textbf{Time (s)}$\downarrow$ & \textbf{SR}$\uparrow$ & \textbf{Time (s)}$\downarrow$ & \textbf{SR}$\uparrow$ & \textbf{Time (s)}$\downarrow$ & \textbf{SR}$\uparrow$ & \textbf{Time (s)}$\downarrow$ & \textbf{SR}$\uparrow$ & \textbf{Time (s)}$\downarrow$ & \textbf{SR}$\uparrow$ & \textbf{Execution Efficiency}$\uparrow$\\
      \midrule
      \textbf{TP-S} & $80\%$ & $\textcolor{Dgreen}{9.5} + \textcolor{Dred}{53.5} = 63.0$ & $40\%$ & $\textcolor{Dgreen}{12.9} + \textcolor{Dred}{90.7} = 103.6$ & $90\%$ & $\textcolor{Dgreen}{8.3} + \textcolor{Dred}{35.6} = 43.9$ & $80\%$ & $\textcolor{Dgreen}{8.2} + \textcolor{Dred}{116.9} = 125.1$ & $100\%$ & $\textcolor{Dgreen}{8.2} + \textcolor{Dred}{55.8} = 64.0$ & $78\%$ & $100.0\%$ \\
      \textbf{TP-D} & $80\%$ & $\textcolor{Dgreen}{8.5} + \textcolor{Dred}{26.1} = 34.6$ & $0\%$ & $Fail$ & $0\%$ & $Fail$ & $0\%$ & $Fail$ & $50\%$ & $\textcolor{Dgreen}{8.3} + \textcolor{Dred}{43.2} = 51.4$ & $26\%$ & $111.4\%$ \\
      \textbf{TwoStep\cite{singh2024twostep}} & $80\%$ & $\textcolor{Dgreen}{11.4} + \textcolor{Dred}{27.3} = 38.7$ & $0\%$ & $Fail$ & $0\%$ & $Fail$ & $0\%$ & $Fail$ & $50\%$ & $\textcolor{Dgreen}{11.3} + \textcolor{Dred}{43.2} = 54.5$ & $26\%$ & $109.2\%$ \\
      \textbf{RoCo\cite{mandi2024roco}} & $80\%$ & $\textcolor{Dgreen}{49.3} + \textcolor{Dred}{26.2} = 75.5$ & $0\%$ & $Fail$ & $90\%$ & $\textcolor{Dgreen}{54.9} + \textcolor{Dred}{24.1} = 79.0$ & $30\%$ & $\textcolor{Dgreen}{155.3} + \textcolor{Dred}{78.9} = 234.2$ & $100\%$ & $\textcolor{Dgreen}{54.4} + \textcolor{Dred}{35.3} = 89.7$ & $60\%$ & $68.7\%$ \\
      \cellcolor{codegray}\textbf{DAG-Plan} & \cellcolor{codegray}$80\%$ & \cellcolor{codegray}$\textcolor{Dgreen}{8.5} + \textcolor{Dred}{27.6} = 36.1$ & \cellcolor{codegray}$50\%$ & \cellcolor{codegray}$\textcolor{Dgreen}{11.7} + \textcolor{Dred}{59.8} = 71.5$ & \cellcolor{codegray}$90\%$ & \cellcolor{codegray}$\textcolor{Dgreen}{8.3} + \textcolor{Dred}{24.0} = 32.3$ & \cellcolor{codegray}$50\%$ & \cellcolor{codegray}$\textcolor{Dgreen}{8.3} + \textcolor{Dred}{69.9} = 77.9$ & \cellcolor{codegray}$100\%$ & \cellcolor{codegray}$\textcolor{Dgreen}{8.3} + \textcolor{Dred}{35.4} = 43.7$ &\cellcolor{codegray}$74\%$ & \cellcolor{codegray}$152.8\%$ \\
      \bottomrule
    \end{tabular}
    }
  \end{center}
\vspace{-10pt}
\end{table*}


This tier evaluates the inherent planning capability of each method, isolating the quality of the high-level plan from low-level execution errors. Our analysis of the plan tests shown in Tab. \ref{plan_test_results} reveals the fundamental advantages of the structured DAG representation. It directly addresses Q1, Q2 and Q3.

\subsubsection{Q1 DAG Representation vs. Linear Sequences}
The DAG-Plan outperforms linear sequence methods. While TP-S is reliable yet inefficient, and TP-D is efficient yet unreliable, DAG-Plan delivers the best of both worlds. DAG-Plan matches the high success rate of TP-S (88\% vs. 92\%) while achieving the stage efficiency of the best dual-arm methods (169.2\%). This shows that the DAG is a necessary representation for reliable and efficient dual-arm planning.

\subsubsection{Q2 Single-Query vs. Iterative Query Cost}
The cost advantage of DAG-Plan's single-query approach is definitive. RoCo's iterative paradigm consumes over 42k tokens per plan, making it prohibitively expensive. DAG-Plan reduces the token cost by nearly $20\times$, achieving a cost profile nearly identical to single-query linear methods but with vastly superior performance.

\subsubsection{Q3 Centralized Graph vs. Decentralized Multi-Agent}
The comparison with Twostep highlights a critical finding: decomposition into fully independent sub-tasks is ill-suited for the intrinsically collaborative nature of a dual-arm system. DAG-Plan's centralized graph provides a global view, enabling it to explicitly model and enforce the necessary interdependencies between arms.

\subsubsection{Attribution of Success and Failure}
A deeper analysis of the planning results reveals the distinct failure modes of each baseline, highlighting the specific advantages of the DAG representation. TP-D fails on complex tasks (Task 2 \& 4) due to the fundamental inability of linear sequences to model non-linear dependencies. Twostep fails as decomposing into independent sub-tasks is ill-suited for interdependent dual-arm coordination. RoCo and DAG-Plan share a common failure ceiling due to the LLM's inherent limitation in temporal dependency prediction. DAG-Plan reaches this LLM-imposed ceiling, having eliminated failures caused by inferior representation TP-D or architecture Twostep.

\subsection{Evaluation with Physical Simulation}
\label{sec:Physical Simulation}

This tier assesses the executability of the generated plans, testing whether high-level logical advantages translate into efficient and feasible physical execution. The simulation results (Tab. \ref{Physical test} and Fig. \ref{demo}) strongly corroborate the planning-level findings and introduce critical insights into physical feasibility. It provides critical evidence that the performance gaps between methods are not just theoretical but manifest in simulation, thereby reinforcing the conclusion of Q1-3 and laying the groundwork for Q4.

\begin{figure*}[htbp]
      \vspace{3pt}
      \centering
      \includegraphics[scale=0.53]{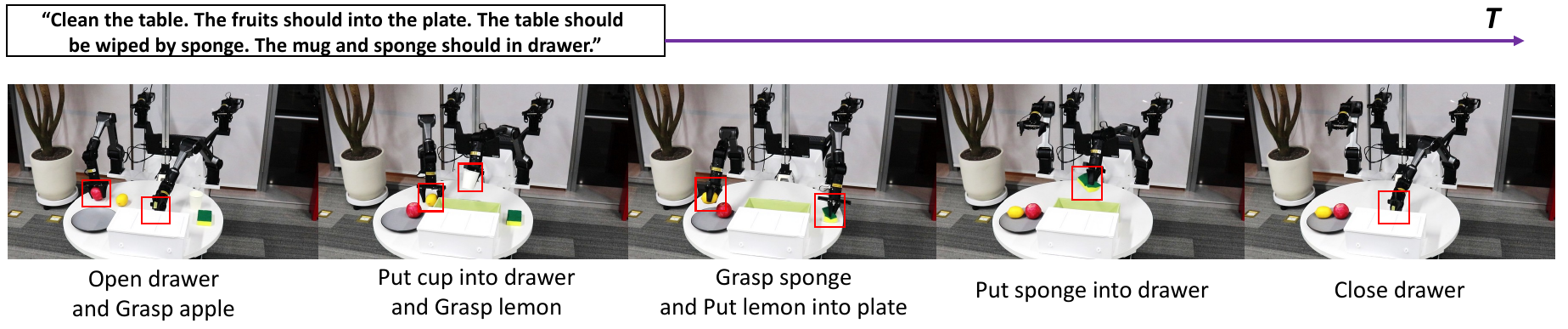}
      \includegraphics[scale=0.53]{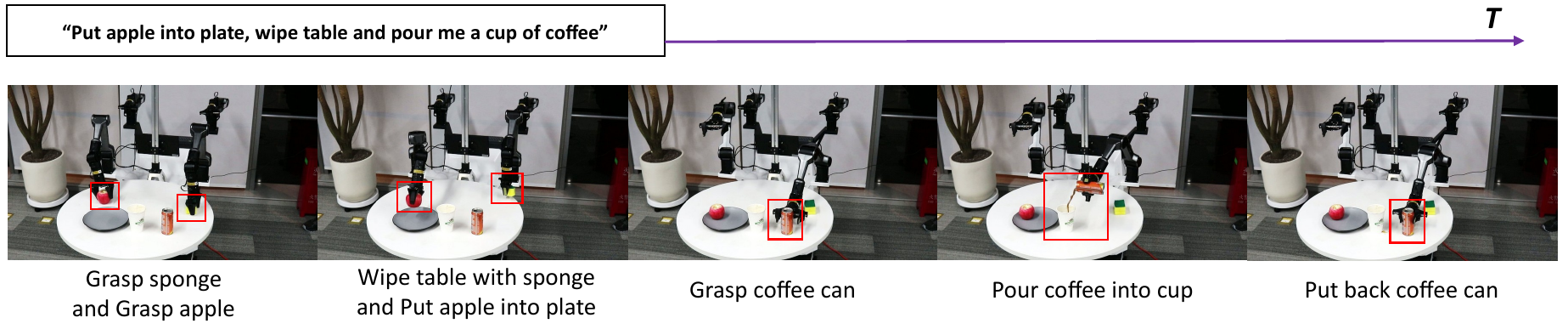}
      \vspace{-10pt}
      \caption{Real-world snapshots of the execution process of long-horizon tasks.}
      \vspace{-13pt}
      \label{realworld demo}
\end{figure*}

\subsubsection{Bridging the Gap from Plan to Action}
A key insight is the stark difference between a logically sound plan and an executable one. TP-D's performance collapses in simulation (26\% SR vs. 40\% in plan tests) because its linear sequences often generate physically infeasible actions (e.g., commanding both arms to grasp adjacent objects, causing collisions). DAG-Plan, equipped with its runtime feasibility checks, dynamically filters out such invalid actions, showcasing its critical role in bridging the sim-to-real gap.

\subsubsection{The Execution-Time Cost of Querying}
The results expose the crippling execution-time cost of the iterative paradigm. RoCo's query time often exceeds its action execution time, leading to an abysmal overall execution efficiency of only 68.7\%. This empirically confirms that query latency is a fundamental bottleneck. DAG-Plan's execution efficiency reached 52.8\% with the single-query approach completely avoids this penalty, achieving a time profile comparable to TP-D but with vastly higher reliability.

\subsection{Real-world Experiments}
\label{sec:Real-world Experiments}
\begin{figure}[htbp]
      \centering
      \includegraphics[scale=0.43]{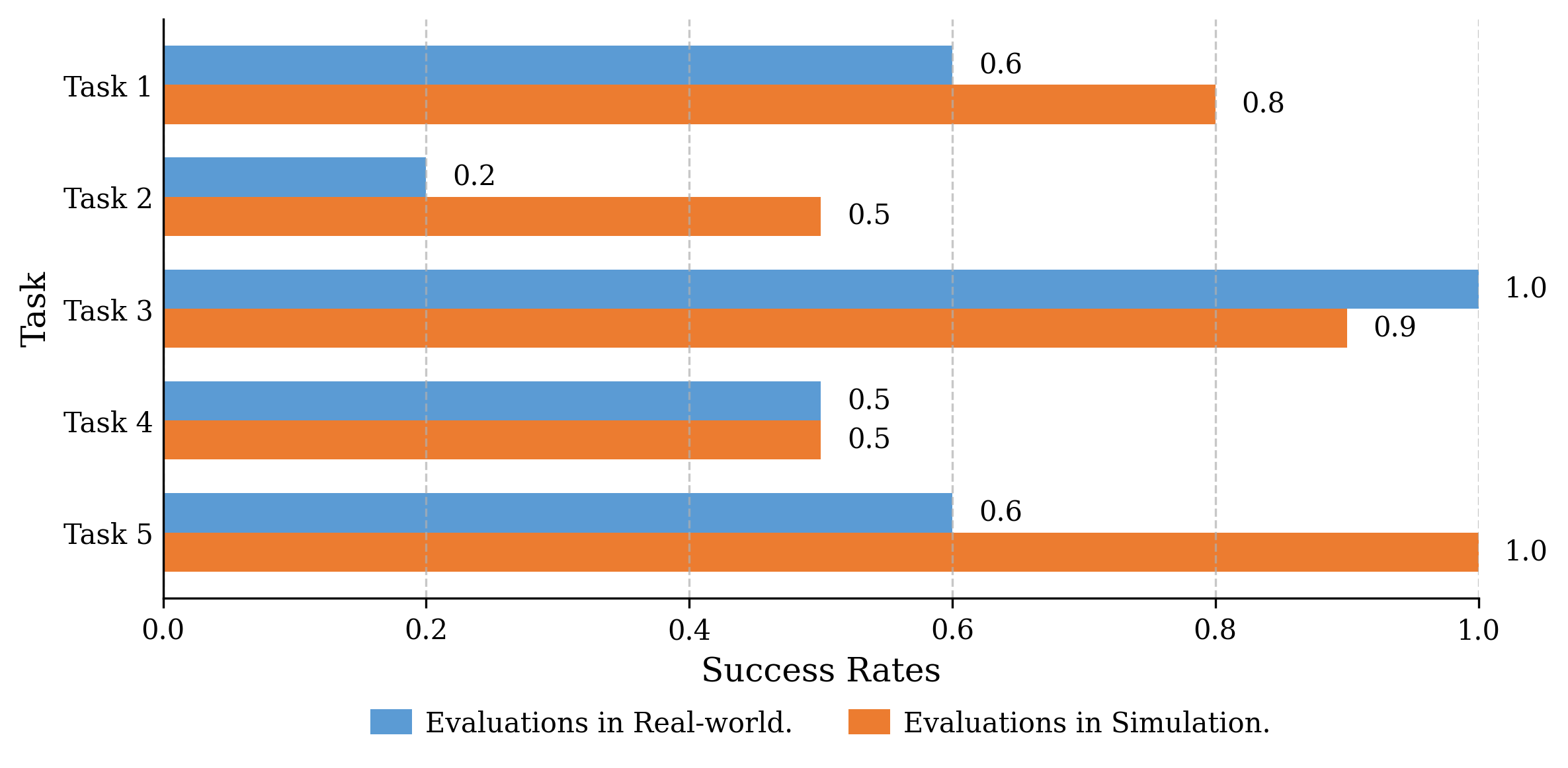}
      \vspace{-20pt}
      \caption{Comparison of real-world and simulation success rates of DAG-Plan.}
      \vspace{-5pt}
      \label{realworld_result}
\end{figure}

We validate DAG-Plan in real-world setting to answer Q4, where a Agilex Cobot Magic dual-arm robot to complete tasks similar to simulation. When dealing with these tasks, the DAG-Plan framework can effectively assign the sub-task for the each arm. We show the execution process of DAG-Plan in real-world in Fig. \ref{realworld demo} first row. We evaluate 10 runs for each task as shown in Fig. \ref{realworld_result} for comparison of real-world and simulation success rates. The slightly lower success rate in real-world is primarily due to the inaccuracy of depth affect the foundation models for execution, whereas the depth in simulations is entirely precise.

In addition to the predefined tasks similar to simulation, we also conducted tests in the wild, as shown in Fig. \ref{realworld demo} second row. By providing human instructions through audio input, the LLM automatically generates DAG, enabling the dual-arm robot to efficiently complete the tasks.

\section{Conclusion}
\label{sec:Conclusion}

This work introduces DAG-Plan, which efficiently and accurately generates collaborative plans with LLMs for dual-arm robots. DAG-Plan decomposes complex tasks into directed acyclic graph (DAG) with clear temporal relationships and iteratively selects feasible sub-tasks based on environmental observations during execution. The main contribution is the replacement of task sequence with a DAG and the dynamic adjustment of the planning according to the current situation, allowing dual-arm robots to flexibly utilize both arms for sub-task execution. Compared to the single-arm approach, DAG-Plan demonstrates significantly higher execution efficiency. In contrast to dual-arm methods, DAG-Plan achieves a higher success rate in task planning. Compared to the iterative method, DAG-Plan maintains the ability to interact with environmental information while requiring shorter query time and reduced token usage.


\bibliographystyle{IEEEtran}
\bibliography{main}{}
 
\end{document}